\documentclass[sigconf]{acmart}
\renewcommand\footnotetextcopyrightpermission[1]{} 
\usepackage{booktabs} 
\usepackage{float}
\usepackage{algorithm}
\usepackage[noend]{algpseudocode}
\usepackage{amsmath}
\usepackage{dblfloatfix}
\usepackage{enumerate}

\setcopyright{rightsretained}

\acmConference[TSMO'18]{Workshop on Two-sided Marketplace Optimization: Search, Pricing, Matching \& Growth}{Feb 2018}{El
  Paso, Texas USA}
\acmYear{2018}
\copyrightyear{2018}

\begin{document}
\title{A Smart Instant Book Filter Model based on Dual-Objective Optimization}

\author{Yi Hou}
\affiliation{%
  \institution{Instant Book}
  \institution{Airbnb Inc}
  \city{San Francisco} 
  \state{California} 
  \postcode{94103}
}
\email{yi.hou@airbnb.com}

\author{Li Fan}
\affiliation{%
  \institution{Instant Book}
  \institution{Airbnb Inc}
  \city{San Francisco} 
  \state{California} 
  \postcode{94103}
}
\email{li.fan@airbnb.com}

\author{Huizhong Duan}
\affiliation{%
  \institution{Search Ranking}
  \institution{Airbnb Inc}
  \city{San Francisco} 
  \state{California} 
  \postcode{94103}
}
\email{hui.duan@airbnb.com}

\renewcommand{\shortauthors}{Y. Hou et al.}

\begin{abstract}
Hospitality service in online marketplaces is becoming more and more popular over the past few years, not only because of its capability to help people easily lease and rent short-term lodging but also it has empowered travellers to connect with locals and feel belonging. A typical pattern of these industries is that there is a central platform mediating the economic interactions between guests and hosts. And the successful booking experience is the key to user satisfaction. However, one of the biggest challenges we are facing is the possible delay of response and chance of rejection when guests try to book a \emph{request-to-book} listing. Previously, we have enabled \emph{instant book} feature to allow hosts to instantly accept qualified guests and guests to confirm booking immediately. Yet the symptom has not been alleviated given mixed \emph{request-to-book} and \emph{instant book} listing supply in search results. One possible solution is to promote all \emph{instant book}  listings in search results however booking rate will be damaged due to some improperly promoted low quality listings. 

In this paper, we present a novel smart \emph{instant book} filter system that applies \emph{instant book} filter intelligently and shifts \emph{request-to-book} to \emph{instant book} without hurting booking rate. Moreover, we reveal the facts how this guided search approach helps land in an improved user experience. Compared with traditional two-sided markets research that targets at finding strategies to maximize a single overall social welfare goal, our proposed system aims to maximize the \emph{instant book} rate with the constraint that booking rate should not be decreased, forming a dual-objective optimization problem.

We show the A/B testing results for our system exposed to global users and the experimental results have demonstrated our model performs better than previous human expert tuned models in terms of both \emph{instant book} rate and overall booking rate. In addition, we've also achieved improved metrics such as the host \emph{instant book} adoption rate. The proposed system has been deployed to production and is serving millions of users everyday.
\end{abstract}

%
%


\maketitle

\section{Introduction}
Recent years' prosperity of industries including Airbnb, Uber, Didi and so on has brought more and more attention to two-sided markets research. The main research themes in this literature are finding strategies to maximize overall social welfare by allocating different resources to valuable agents normally in forms of monetary incentives. One typical approach is by subsidizing one side using customized prices to increase the overall welfare by looking at performance indicators for example the price demand curves, where the prices are normally bounded by marginal cost and agent's willing to pay\cite{Clarke1971, DBLP:journals/jam/Xu13b}. Another track of work is focused on auction based optimization mechanism in two sided markets, where both sides in platform care about the number of existence of the other and platform aims at maximizing the revenue based on two sides' economic interactions\cite{Roson2005, DBLP:journals/corr/abs-1111-0046}. Because of the complexity of real life applications, we are often faced with multiple hard constraints under the optimization model. An interesting track of work by Liu et.al. has attempted to optimize the water-right market with the constraint that a minimum transaction threshold must be enforced to offset administrative and other costs\cite{conf/aaai/LiuTXZ16}, thus the transaction must either be 0 or above the threshold. 

While there is a rich literature studying various problems in two-sided markets, previous approaches are not feasible to our problem setting where we want to convert \emph{request-to-book} to \emph{instant book} without hurting booking rate. In details, the tradeoff we are dealing with is to promote \emph{instant book} without cutting supply much or nudging searchers away from great supply which is almost but not \emph{instant book}. 

More specifically, we investigate and differentiate our problem from two perspectives:  

\textbf{Our problem is not deterministic}: once smart \emph{instant book} filter is applied for a specific search request, there is no direct rule or ground truth to indicate if a booking transaction will be made afterwards, which increases the difficulty for modeling and evaluation.

\textbf{Our problem has dual-objectives}: instead of optimizing one goal, our smart \emph{instant book} filter system needs to maximize the \emph{instant book} rate at the same time keep booking rate positive or neutral. 

In this work, we show a novel smart \emph{instant book} filter system that addresses these challenges. To address the first line, while we don't have a direct rule to indicate if a booking transaction will be made after filter, we leverage the discounted cumulative gain as a function proxy to measure the ranking quality and booking desirability given it is a standard measurement of search engine algorithms\cite{Jarvelin:2002:CGE:582415.582418}. Typically this discounted cumulative gain considers a rich set of attributes including guest host matching score, listing average review rating, ranking position and more to model possibility of booking. Using discounted cumulative gain as a function proxy, the model learning process transforms to explore parameters that maximize \emph{instant book} discounted cumulative gain inside search results while keeping overall booking discounted cumulative gain change positive or neutral. To answer the second line, we adopt random walk parameter exploration with restart to human expert picks. In each learning episode, we start from human expert picked parameters and randomly extend to all directions. This learning episode keeps on as long as both \emph{instant book} discounted cumulative gain and booking discounted cumulative gain changes are positive when compared to production search which doesn't enable smart \emph{instant book} filter system. The episode stops when booking discounted cumulative gain change falls to negative and we track the best parameters we've observed so far during the learning episode. We monitor discounted cumulative gain changes via our in house side by side ranking simulation system. Additionally, we iterate multiple learning episodes with different restarts to avoid falling into local optima or overfitting. We'll elaborate more in Section 3 about learning algorithm and test the validity of our function proxy in experiments under Section 4. Furthermore, by analyzing the experiment results, in addition to the targeted \emph{instant book} rate increase, we also find some interesting second order effect about host \emph{instant book} adoption rate gain, which is comparable to strong host-side \emph{instant book} adoption campaigns. This signifies that the interaction between guests and hosts is remarkably important, which paves the way for more advances in this area.

The rest of this paper is organized as follows: Section 2 introduces the preliminaries of the system, Section 3 describes our model and algorithm, Section 4 shows the experiment results using real data, Section 5 illustrates the overall system and user experience and lastly we conclude our findings and shed light on future work in Section 6.

\section{Preliminaries}
\subsection{Instant Book and Smart Instant Book Filter}
People like it when things are fast. We want our deliveries within an hour, our downloads within a minute, and our news in an instant. That is why we need to help our Airbnb community to book listings they want instantly. \emph{Instant book} is a key lever in our product to empower the Airbnb community to book places immediately without prior host approval and increase acceptance. Before we offered \emph{instant book}, guests using reservation requests can often get multiple rejections and the situation may turn out worse for travelers who want to make their plans quickly. With \emph{instant book}, guests can filter on our search page to find places where reservations can be confirmed instantly. Still, we've realized some guests select \emph{request-to-book} listings without knowing the chance of rejection and possible delay in response. Motived by this, we've initiated the smart \emph{instant book} filter system---intelligently applying the \emph{instant book} filter to searches to encourage a better user experience. Our landscape of \emph{instant book} has changed incredibly over the past few years and today 2 out of 3 booking are made using \emph{instant book}. Since 2016, we have been steadily improving \emph{instant book}, and one of the biggest levers is smart \emph{instant book} filter.

\subsection{Background}
In this section, we first describe some definitions and notations that we use throughout this paper. Then we formally describe the problem we are solving. 

\textbf{Book, instant book and request-to-book}: guests can book a listing either by \emph{instant book} or \emph{request-to-book}, where the former can confirm booking instantly and the latter requires host's manual approval. The total number of booking transactions made over a certain period of time is the summation of  \emph{instant book} and \emph{request-to-book} bookings. In our experiment settings, all these quantities are accumulated over the complete experiment period.

\textbf{Discounted cumulative gain}: generally discounted cumulative gain is a standard measurement of ranking quality. It is the utility function of interest discounted by intertemporal effects or distance to present status. More formally, we adopt the standard discounted cumulative gain curve thus we have 
$DCG_p= \sum_{i = 1}^{p} \frac{2^{rel_i} - 1}{log_2 (i + 1)}$, where $rel_i$ is the projected relevance of the result at position i and p is the particular rank position we consider. In our application, p represents the set of search result listings we take into consideration when computing utility value. In our experiment, we consider the first page of search result listings for p. $rel_i$ is a trained value supplied in ranking model considering many attributes including listing average review rating, price and so on. Booking discounted cumulative gain $U_B$ now becomes the discounted cumulative gain over the first p result listings in search page and \emph{instant book} discounted cumulative gain $U_I$ is the discounted cumulative gain value for all the \emph{instant book} listings among the p results. In our experiment settings, p is 18 given it is the number of total listings in the first search page.

\textbf{Information vector $w$}: information vector, annotated as $\omega$ is a vector of informative attributes that are available in search backend, including but not limited to number of listings available for this search, number of \emph{instant book} listings available for this search, overall discounted cumulative gain, \emph{instant book} discounted cumulative gain. 

\textbf{Branching value vector $v$}: $v$ is a vector of threshold values that has the same length of information vector $w$, where we can perform 1:1 comparison between values in each vector to decide if all conditions satisfy.

 \textbf{Branching logic $\phi$}: $\phi$ is a high level rule that defines how we compare information vector $w$ with branching value vector $v$, one naive example is that $\phi$ returns true if and only if for every vector index, value in $w$ is larger than value in $v$. 

\textbf{Learning Episode}: a learning episode represents a continuous process that starting from human expert picked parameters, we randomly explore in parameter space to find best parameter values for our model.

\textbf{Search Iteration}: a search iteration is a complete search process that contains fixed search query and response.

\textbf{API server}: the server that accepts different client's API requests (i.e. from web or mobile) and bridges the communication between client and backend service.

\subsection{Problem formation}
With these concepts in mind, we next formally model the problem as a bipartite graph $\mathcal{G} = (N, E)$. Here $N = U \cup L $ where U is the set of guests and L is the set of listings owned by hosts, thus N represents the union of guests and listings. Each edge $e_{ij}^{o} \in E$ represents a booking transaction between a guest and listing and the type $o \in \{0, 1\}$ stands for the type of booking, where 0 represents \emph{request-to-book} and 1 represents \emph{instant book}. Every edge has an equal unit weight. Fig 1 illustrates the formalized problem. 

Therefore the goal here is to maximize the value of $\sum e_{ij}^{1} \in E$ as well as to keep $\sum e_{ij}^{o} \in E$ where $o \in \{0, 1\}$ as good as production without smart \emph{instant book} filter's status. Considering the drawback that once smart \emph{instant book} filter is applied as we shift \emph{request-to-book} to \emph{instant book} there is no direct rule to infer the existence of each edge, so instead of exactly measuring the quantity of edges, we adopt discounted cumulative gain as a function proxy to reflect these quantities' distribution changes. Since discounted cumulative gain is mapped 1:1 to each individual search iteration, thus we need to perform multiple searche iterations and average the discounted cumulative gain changes to converge to real booking quantity percentage change. Afterwards, the problem transforms to maximize the average \emph{instant book} discounted cumulative gain while not decreasing average booking discounted cumulative gain. We verify the validity of discounted cumulative gain proxy in experiment part under Section 4 and experiment result proves that the real booking and \emph{instant book} quantities change monotonically with the averaged booking and \emph{instant book} discounted cumulative gain change.

\begin{figure}[h]
\centering
\includegraphics[height=2.2in, width=0.5\textwidth]{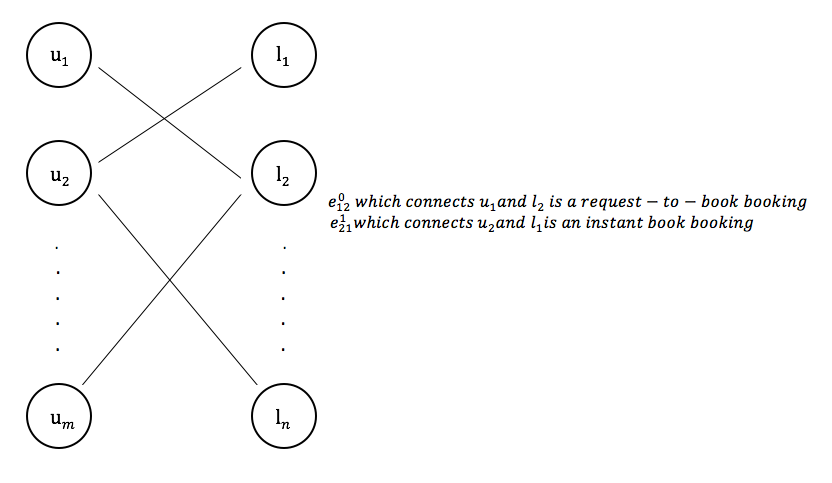}
\caption{Graph representation of booking entities and relationship}
\end{figure}

\section{Method and Optimization}
In order to solve the problem above, we need to find a model $DT(\omega)$ that based on the information vector $\omega$ available in search backend as we aggregate search results, decides if the smart \emph{instant book} filter shall trigger. More formally, we need to construct $DT(\omega)$ that:

\begin{equation}
\begin{aligned}
& {\text{maximize}} & & \frac{1}{K} \sum_1^K  U_I \\
& \text{subject to} & & \sum_1^K  \Delta U_B \ge 0, \\
& & & where \; K \; is \; a \; fixed \; number \; of \; search \; iterations.
\end{aligned}
\end{equation}

Let's recall again that $U_B$ is booking discounted cumulative gain and $U_I$ is \emph{instant book} discounted cumulative gain associated with each search iteration respectively. 

We adopt deterministic decision tree as the underline model of $DT(\omega)$ because of its high interpretability. We implement branching logic $\phi$ inside decision tree that given the information vector $\omega$ and branching value vector $v$, it outputs a boolean flag indicating if smart \emph{instant book} should trigger. Then, our core task becomes to find a vector of branching values $v$ such that $\phi(v, w)$ maps to false or true and branching logic $\phi$ checks $w$ with our branching values $v$. Please note that our learning problem here is not the same as traditional decision tree threshold learning problem on continuous variables for classification, because: 1) we don't have any labeled data for training and evaluation and 2) our core metrics are not accuracy/precision but average discounted cumulative gain values. Therefore we can't use traditional tree learning approaches in this scenario. Next we formally introduce the optimization system.

\subsection{Optimization System}
To solve equation (1), we adopt random walk parameter exploration with restart to human expert picks, where in each step of learning episode, we randomly pick a parameter in branching value vector $v$ and either increase it or decrease it. Then we use side by side ranker (SxS ranker), which is our in house ranking simulation system, to verify the average discounted cumulative gain change. In our experiment, we average discounted cumulative gain change on $K=$ 10,000 search iterations for each episode step. Fig 2 shows one complete learning episode.

\begin{figure}[h]
\centering
\includegraphics[height=2.2in, width=0.5\textwidth]{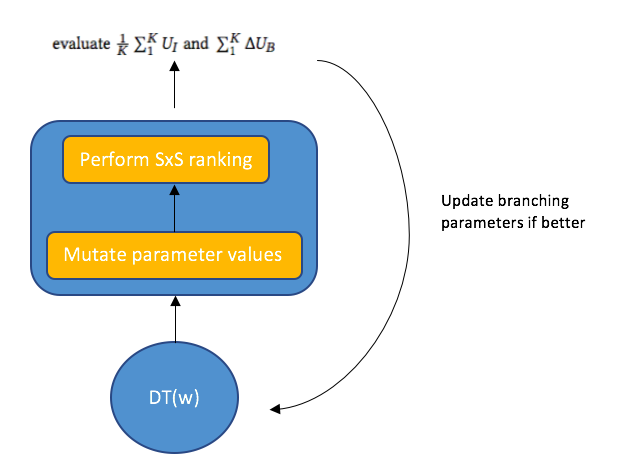}
\caption{One complete learning episode}
\end{figure}

\subsection{Algorithm}
Between multiple learning episodes, we keep track of the best performance so far in terms of $v$ and $\Delta U_I $. In the end, we use the best $v$ values we observe in parameter learning process as our experiment values for production A/B testing. Formally, we define the optimization algorithm in table Algorithm 1.

\makeatletter
\def\BState{\State\hskip-\ALG@thistlm}
\makeatother

\begin{algorithm}
\caption{Parameter learning process}\label{euclid}
\begin{algorithmic}[1]
\Procedure{learning}{}
\State $\textit{v} \gets \text{human expert picked values}$
\State $\text{maximum} \; \Delta U_I \gets \textit{0}$
\BState \emph{top}:
\If {$ \Delta U_I \text{ is good enough}$} \Return {$v$}
\EndIf
\BState \emph{loop}:
\State $\text{Mutate a value in } v$.
\State \emph{loop K search queries}:
\If {$ \Delta U_B \geq 0 $}
\State $\text{Update best } v \text{  and best } \Delta U_I$.
\Else
\State \textbf{goto} \emph{top}.
\EndIf
\EndProcedure
\end{algorithmic}
\end{algorithm}

\begin{figure*}[t]
\centering
\includegraphics[height=0.6\textwidth, width=0.9\textwidth]{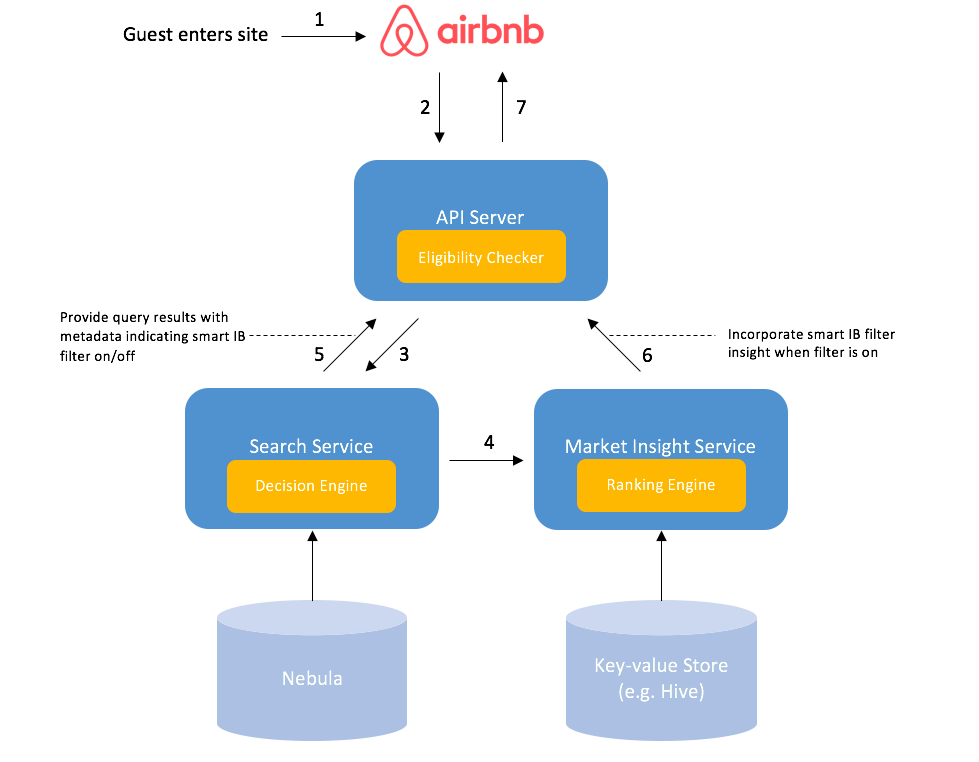}
\caption{Smart Instant Book filter system workflow}
\end{figure*}
In general, we first initiate with human expert selected branching vector $v$. Then we try to mutate one value in $v$ and perform side by side ranking over 10,000 search iterations to see the \emph{instant book} discounted cumulative gain change and booking discounted cumulative gain change. We keep track of the best thresholds we've observed so far and stop episode when we see a negative booking discounted cumulative gain change. After each learning episode, we manually evaluate the \emph{instant book} discounted cumulative gain change to make a decision if that is significant enough. Next we illustrate how our learned parameters work in real traffic experiments.

\section{Experiment}
\begin{table*}[t]
  \caption{Experiment results comparing our learned parameters with human expert picked parameters}
  \label{tab:freq}
  \begin{tabular}{ccl}
    \toprule
    Parameters & Instant Bookings Gain & Bookings Gain\\
    \midrule
    Human expert picked parameter set 1 &  11\% & -0.90\%\\
    Human expert picked parameter set 2 & 3.50\% & -0.80\%\\
    Human expert picked parameter set 3 & 26\% & -1.60\%\\
    Human expert picked parameter set 4 & 21\% & -1.60\%\\
    Human expert picked parameter set 5 & 15\% & -1.30\%\\
    Human expert picked parameter set 6 & 6.10\% & -0.17\%\\
    Learned parameter set & 8.20\% & 0.00\%\\
  \bottomrule
\end{tabular}
\end{table*}

\begin{figure}[h]
\centering
\includegraphics[height=0.3\textwidth, width=0.5\textwidth]{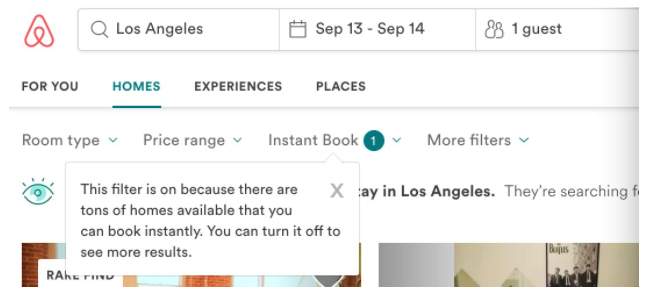}
\caption{Smart instant book filter with its tooltip}
\end{figure}

\begin{figure}[h]
\centering
\includegraphics[height=0.4\textwidth, width=0.5\textwidth]{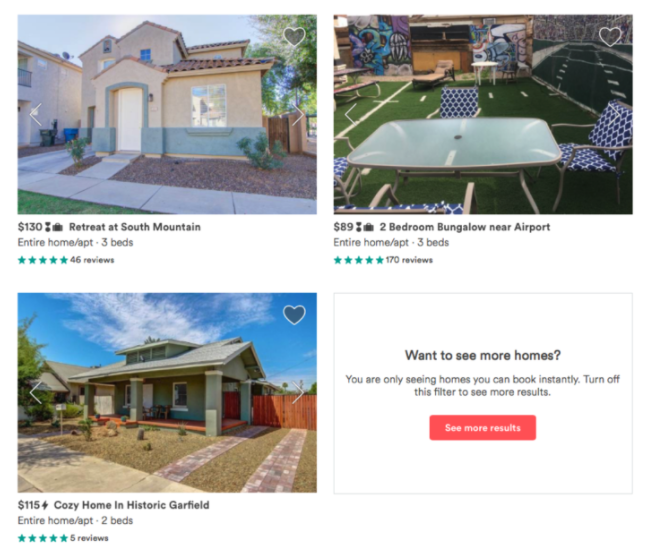}
\caption{Contextual market insight reminder card}
\end{figure}

In this section, we show some preliminary results in Table 1 as we perform A/B testing using the branching value vector $v$ we have learned in Section 3 in real global traffic. We demonstrate that our result has outperformed all previous six sets of human expert picked thresholds. In addition, we reveal some interesting second order effect associated with this experiment.

Generally, as Table 1 illustrates, only the model using our learned branching value vector has successfully achieved the dual-objectives that: 1) to significantly increase \emph{instant bookings} and 2) at the same time to keep bookings not decreased. Though we can observe that some human expert picked parameters have an even bigger \emph{instant books} gain, yet the total bookings' decrease is not affordable in the problem settings.

Moreover, we observe that \emph{instant books} and bookings change proportionally with \emph{instant book} discounted cumulative gain and booking discounted cumulative gain. When we compare our offline analysis with production experiment data report, we've found that the real booking and \emph{instant book} percentage changes are just values by applying constant factors to discounted cumulative gain changes. This validates the correctness of using discounted cumulative gain as a function proxy to represent booking distribution. A second order effect we have identified in experiment is the tendency of hosts to adopt \emph{instant book} for their listings, as in our experiment $+0.96\%$ hosts adopt \emph{instant book}, which is comparable to strong host-side \emph{instant book} adoption campaigns. The signifies the boundary between guests and hosts is blurry---many guests are hosts as well and the interaction between guests and hosts is remarkably important, which paves the way for more advances in this area.

\section{System and user experience}
After introduction of the core model and learning algorithm, we now explain how our learned model fits in the overall system here. Our learned model is the key part embedded in the decision engine inside search service, where \emph{instant book} filter triggering logic is computed. The high level architecture is shown in Fig 3. When a guest uses Airbnb to search for homes, the API server first collects basic query information and performs an initial eligibility check. Next, the API server queries the search service and talks with the market insights service\footnote{Market insight is our service that surfaces personalized market data to end users.} to decide if the \emph{instant book} filter should be triggered, generating corresponding candidate market insights in real-time. Finally, the API server combines the information collected from both services and reflects the result in the user interface.

More specifically, the workflow is as follows:

\begin{enumerate}[Step 1:]
\item User first lands on our platform.
\item User searches for homes via the search box. API server validates basic information in this phase for example whether the query contains check-in and check-out dates to flag queries that are eligible for smart \emph{instant book} filter.
\item API server sends a search request with smart \emph{instant book} filter eligibility flag to search service, which talks with Nebula---our in house storage platform, to perform ranking and generate search results. Our decision engine sits inside search service. It first aggregates search results, then it generates information vector $w$ and finally uses the deterministic decision tree model $DT(w)$ to decide whether to trigger smart \emph{instant book} filter. 
\item Search service passes market availability metadata to market insight service where candidate insights including the smart \emph{instant book} filter insight will be ranked based on their values to the guest.
\item Search service returns results to server.
\item Market insight service returns final eligible insights to server.
\item Api server combines all information and powers the front-end to display.
\end{enumerate}

Fig 4 is the product experience when user's search triggered smart \emph{instant book} filter, which comes together with a tooltip to alert users the filter is on. Moreover, if a user keeps zooming in the map to narrow down searching area while \emph{instant book} filter is on, which consequently find fewer and fewer \emph{instant book} results, a market insight card will replace a single listing card to remind guests they can release smart \emph{instant book} filter with one click. The market insight card is shown in Fig 5.

\section{Conclusions and Future Work}
In our model, we've used deterministic decision tree with fixed parameters. As we are iterating on each learning episode, we've observed interesting tradeoffs through trees. For instance, when we are reducing the minimum \emph{instant book} listings number requirement in the branching value vector, the \emph{instant book} discounted cumulative gain increases but the overall booking discounted cumulative gain first increases and then decreases. This agrees with our intuition that by gradually lowering the minimum \emph{instant book} listings number requirement for smart \emph{instant book} filter, we should see \emph{instant book} discounted cumulative gain increase since more searches are eligible for smart \emph{instant book} filter. However, when we are over promoting \emph{instant book} listings at certain point and end up with some low quality listings promoted, as a consequence, this will reduce the overall booking discounted cumulative gain.

As we are constantly iterating on smart \emph{instant book} filter system, in addition to the current state of model, we are planning to leverage automated machine learning infrastructure and integrate system with an online scoring model in our next step to capture more evolving conditions. This will open two directions: explore other parameter optimization approaches and different models.

Previously we've dedicated tremendous amount of effort to host side projects and this project is where we started on guest side and landed with good results. As a result, our work has driven \emph{instant book} percentage by more than $5\%$, blending in a substantial way of company growth. 


\begin{acks}
  The authors would like to thank Surabhi Gupta, Ricardo Bion, Trunal Bhanse, Lenny Rachitsky, Brendan Collins, Martin Nguyen, Peng Dai and Jeff Fossett for their support and generous help in many ways.

  The authors would also like to thank the anonymous referees for
  their valuable comments and helpful suggestions. 
\end{acks}

\bibliographystyle{ACM-Reference-Format}
\bibliography{sample-bibliography} 

\end{document}